\documentclass[12pt]{iopart}
\usepackage{rotating}
\usepackage{color}
\usepackage{caption,subcaption}
\usepackage{multirow}
\usepackage{booktabs}
\usepackage{tabularx}
\usepackage{enumitem}
\usepackage{threeparttable}

\usepackage[breaklinks,colorlinks = true,linkcolor=blue,urlcolor  = blue,citecolor = blue,anchorcolor = blue]{hyperref}

\usepackage{iopams} 
\usepackage[normalem]{ulem}
\usepackage[mathlines,displaymath,running]{lineno}
\usepackage{xcolor}
\renewcommand{\figurename}{Figure}

\renewcommand{\today}{\ifcase \month \or January\or February\or March\or %
April\or May \or June\or July\or August\or September\or October\or November\or %
December\fi \ \number \year} 

\begin{document}
\title[]
{Few-Shot Transfer Learning for Individualized Braking Intent Detection on Neuromorphic Hardware}

\author{Nathan A. Lutes$^1$, Venkata Sriram Siddhardh Nadendla$^{2,*}$, \\
K. Krishnamurthy$^1$}

\address{$^1$ Department of Mechanical and Aerospace Engineering,
Missouri University of Science and Technology, 400 W. 13th Street, Rolla, MO, USA, 65409}

\address{$^2$ Department of Computer Science, Missouri University of Science and Technology, 500 W. 15th Street, Rolla, MO, USA, 65409.}

\address{$^*$ Author to whom any correspondence should be addressed.}

\ead{nalmrb@umsystem.edu, nadendla@mst.edu, kkrishna@mst.edu}

\vspace{10pt}
\begin{indented}
\item[]\today
\end{indented}

\begin{abstract}
\emph{Objective}: This work explores use of a few-shot transfer learning method to train and implement a convolutional spiking neural network (CSNN) on a BrainChip Akida AKD1000 neuromorphic system-on-chip for developing individual-level, instead of traditionally used group-level, models using electroencephalographic data. The efficacy of the method is studied on an advanced driver assist system related task of predicting braking intention. 
\emph{Approach}: Data are collected from participants operating an NVIDIA JetBot on a testbed simulating urban streets for three different scenarios. Participants receive a braking indicator in the form of: 1) an audio countdown in a nominal baseline, stress-free environment; 2) an audio countdown in an environment with added elements of physical fatigue and active cognitive distraction; 3) a visual cue given through stoplights in a stress-free environment. These datasets are then used to develop individual-level models from group-level models using a few-shot transfer learning method, which involves: 1) creating a group-level model by training a CNN on group-level data followed by quantization and recouping any performance loss using quantization-aware retraining; 2) converting the CNN to be compatible with Akida AKD1000 processor; and 3) training the final decision layer on individual-level data subsets to create individual-customized models using an online Akida edge-learning algorithm.
\emph{Main Results}: Efficacy of the above methodology to develop individual-specific braking intention predictive models by rapidly adapting the group-level model in as few as three training epochs while achieving at least 90\% accuracy, true positive rate and true negative rate is presented. Further, results show the energy-efficiency of the neuromorphic hardware through a power reduction of over 97\% with only a $1.3\times$ increase in latency when using the Akida AKD1000 processor for network inference compared to an Intel Xeon central processing unit. Similar results were obtained in a subsequent ablation study using a subset of five out of 19 channels. 
\emph{Significance}: Especially relevant to real-time applications, this work presents an energy-efficient, few-shot transfer learning method that is implemented on a neuromorphic processor capable of training a CSNN as new data becomes available, operating conditions change, or to customize group-level models to yield personalized models unique to each individual.

\end{abstract}

\section{Introduction}
Research in computing hardware, including graphics processing units (GPUs) and field programmable gate arrays, has rapidly accelerated in recent years, routinely producing faster and more powerful processing units. However, von-Neumann architecture hardware suffers from a notoriously stubborn limitation - tremendous computing power is achieved at the cost of higher energy requirements. This has led to a more focused examination of alternative hardware paradigms that can approach the capabilities of state-of-the-art central processing units (CPUs) and GPUs, but with significant energy savings. One promising approach is \emph{neuromorphic hardware}, a new computing architecture that closely resembles biological neural networks in its construction and operation \cite{Schuman_survey_2017,Clark_Wu_2023,Shrestha_Fang_Mei_Rider_Wu_Qiu_2022}. These types of processors use asynchronous, small bit-width encoded message passing, similar to the brain, to achieve their remarkable energy efficiency. The most recent advances in neuromorphic computing have shown competitive latency at drastically less energy consumption than the best von-Neumann style devices \cite{Davies2021AdvancingNC}. The major reduction in energy required for inference and online training of algorithms lends neuromorphic processors to be prime candidates for use in shared-resource environments; especially, in power-constrained applications.

Alongside the technical advancement in the design and development of brain-inspired neuromorphic hardware, the design and development of biologically influenced learning algorithms has also progressed \cite{Shrestha_Fang_Mei_Rider_Wu_Qiu_2022,Schuman_Kulkarni_Parsa_Mitchell_Date_Kay_2022}. One such example is the so-called `third’ generation of neural networks - spiking neural networks (SNNs) \cite{Tan_Sarlija_Kasabov_2020,Dora_Kasabov_2021,Tavanaei_Ghodrati_Kheradpisheh_Masquelier_Maida_2019,Maass_1997}, including deep learning versions of SNNs, such as convolutional spiking neural networks (CSNNs). Sharing in similarity to how neuromorphic hardware takes inspiration from biological neural networks, SNNs also mimic these networks in the manner in which they pass information from one layer to the next. The output of each neuron is a vector of binary values along a temporal dimension with the ``on" (i.e., 1) values being referred to as ``spikes". Most spiking neurons contain internal dynamics that transform the weighted sum of input spikes from the previous layer into a retained memory state. When this state reaches a critical threshold, the neuron emits a spike and the state experiences a total or partial reset. This mechanism dictates the frequency of the output spikes over time with the output of the neuron being ``off" (i.e., 0) for every internal state value below the critical threshold. Although several experiments have been used to demonstrate comparable performance of SNNs with that of state-of-the-art artificial neural networks \cite{Christensen_2022}, their main appeal lies in their inherent ability to be directly mapped to neuromorphic hardware allowing straightforward exploitation of the processor’s energy efficiency attribute. However, a limitation to exploit the full potential of SNNs has been the slow advancement in the design of on-chip training algorithms, which is being addressed with the recent availability of an edge learning algorithm integrated within the Akida AKD1000 processor \cite{Vanarse_Osseiran_Rassau_van_der_Made_2019,Ferré_Mamalet_Thorpe_2018}.

\subsection{EEG-Based Advanced Driver Assistance}
One technical field that could potentially benefit from adoption of neuromorphic hardware and SNNs is wireless brain-computer-interface (BCI) technologies \cite{doubi_survey}. Electroencephalograph (EEG), commonly associated with BCI, is increasing in popularity because of its non-invasiveness, mobility and flexibility, making it ideally suited for practical applications beyond clinical settings. One emerging sub-field of BCI is EEG-based advanced driver-assist systems (ADAS), a group of assistive technologies designed to assist with driving and/or parking decisions thereby aiding a driver in safely operating their vehicle \cite{8620826}. An example of an EEG-ADAS system is described in Ford Global Technologies patent for a vehicle control system using electrical impulses from motor cortex activity in the driver's brain \cite{ford_patent}. This system is used to help train a world champion cyclist to become a rally car driver by monitoring his EEG signals to better understand the driver's state during training \cite{EEG_racer}.

Indeed, the field of intelligent ADAS, or ADAS systems imbued with computer-driven decision making and control capabilities, has seen progress in recent years, including some early work involving neuromorphic computing techniques. For example, a NM500 neuromorphic processor that contains 76 neurons was used to recognize traffic information images marked on the road (such as crosswalks) and output a speed control signal, thus creating an effective vehicle speed regulation system \cite{10.1007/978-3-030-14907-9_1}. In another work, the NM500 processor was used for pedestrian image detection on an embedded device, and its performance and energy savings are compared against pedestrian detectors designed for CPU and GPU hardware \cite{electronics9071069}. Akida neuromorphic system-on-chip (NSoC) has also been deployed for ADAS related purposes \cite{brainchip_design_smarter_cars,Ward-Foxton_2022b}; however, this did not include specific EEG-ADAS applications. In light of the growing interest in the use of neuromorphic hardware and ADAS, and noting the relative infancy of these two fields, there exists numerous opportunities for significant progress to be made.

\subsection{Relevant Work}
Related to developing individual-level models, meta-learning (a.k.a. ``learning to learn") has been presented as an approach to optimize the learning process itself, instead of optimizing a single model for a single task \cite{thrun_pratt_1998,andrychowicz_2016,hochreiter_younger_conwell_2001}. ``One-shot learning", an extreme sub-method of meta-learning, has been used previously
to train a SNN 
in only ``one-shot" or by only showing the network an input associated with the new task one time to produce sufficient learning \cite{scherr_stockl_maass_2020}. This was accomplished by augmenting SNNs with adaptive neurons to increase their ability to generalize and by using two neural networks, one for a ``teacher" and another for a ``learner"; the learner being responsible for actually learning any new task, and the teacher learning an optimal learning signal to distribute to the learner such that its learning of the new task is optimized. The learning optimization, in this case, is accomplished through a gradient estimation strategy referred to as ``natural e-prop". In another approach, using entropy theory, a gradient-based few-shot learning scheme in a recurrent SNN architecture was presented \cite{Yang_Linares-Barranco_Chen_2022}. The potential energy efficiency of implementing this scheme in a neuromorphic system was discussed, but no actual results were presented. Few-shot transfer learning was previously explored through development of a ``meta-transfer learning" method which combines the training schemes of few-shot learning and transfer learning to more effectively utilize typical transfer learning models for few-shot learning \cite{Sun_Liu_Chua_Schiele_2019}. However, their work did not include expansion to deep spiking neural networks or deployment to real hardware.

Although previous studies detailing neuromorphic hardware within an ADAS context is limited, studies utilizing EEG-adjacent data, such as EMG data, have demonstrated the capability and utility of neuromorphic hardware. For example, neuromorphic hardware was used in the classification of hand signals \cite{Donati_Payvand_Risi_Krause_Indiveri_2019}, and in an event-based camera sensor fusion approach \cite{Ceolini_Frenkel_Shrestha_Taverni_Khacef_Payvand_Donati_2020}. The former study included a rigorous analysis of the hardware during the analysis task and obtained classification accuracy greater than 80\% with hardware power consumption of only 0.05 mW. The latter study expands the EMG-only input to also include an event-based camera and compares a fully neuromorphic processing approach to a traditional machine learning technique on a portable GPU. The authors report similar discrimination performance between the two systems with the neuromorphic hardware showing increased latency but substantial improvement in energy efficiency.

In a previous study by the current authors, EEG data was used to predict prior-to-movement driver braking intention using biological systems' inspired algorithms for the first time \cite{Lutes_Nadendla_Krishnamurthy_2024}. The CSNN algorithm was chosen because of its compatibility with energy-efficient neuromorphic hardware, but the actual implementation was not considered. Further, this prior work was focused on developing a group-level model using data obtained from participants in a stress-free environment.

\subsection{Contribution}
The main contribution of this work is to present an energy efficient, few-shot transfer learning method for training a deep SNN using BrainChip Akida AKD1000 processor for development of individual-level cognitive models and demonstrate the efficacy of this method on an EEG-based ADAS application to detect anticipatory brain potentials at the individual level. Past EEG studies focused on group-level models aimed at generalizing to a large population. However, this scale of model can be less than ideal because individuals exhibit different responses to the same task. Therefore, it is imperative to develop and use models customized to each individual, especially in real-time applications. The use of a neuromorphic processor is also novel in the context of EEG-ADAS literature. 

This work extends the previous results of \cite{Lutes_Nadendla_Krishnamurthy_2024} on predicting braking intention on two fronts: i) individual-level models are developed using a few-shot transfer learning method; and ii) two additional experiments are considered: one with elements of physical fatigue and cognitive distraction \cite{Xu_Zhang_He_Zhao_Qi_Zhou_Zhang_Ming_2018,Almahasneh_Chooi_Kamel_Malik_2014} and the other with pseudo-realistic braking stimulation in the form of traffic light signals.

\subsection{Organization}
Details of the EEG experiments, EEG data preprocessing and Akida AKD1000 processor details can be found in Section \ref{preliminaries}. The few-shot transfer learning methodology and its application to the EEG datasets are explained in Section \ref{method}, and results are presented in Section \ref{results} and discussed in Section \ref{discussion}. Finally, the paper concludes in Section \ref{conclusion}.

\section{Experiments, Datasets and Hardware}\label{preliminaries}

\subsection{Experiment Design and Data Acquisition}\label{data acquisition}

\begin{figure}
    \centering
    \includegraphics[width=0.5\linewidth]{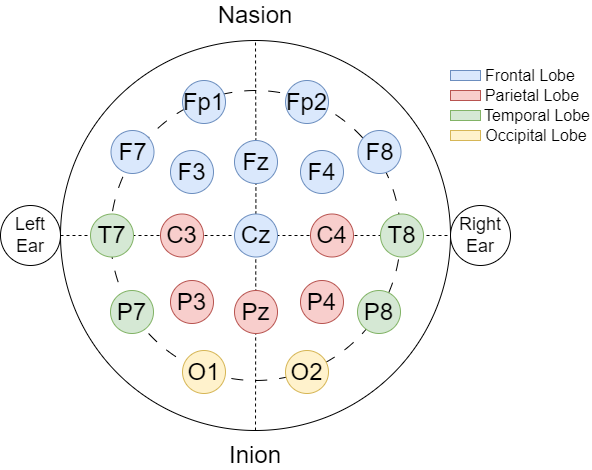}
    \caption{Schematic of the 10-20 configuration of EEG electrodes used for collecting data in this study. Grouping of the EEG electrodes correspond to frontal, parietal, temporal and occipital brain lobes \cite{Rojas_Alvarez_Montoya_de_la_Iglesia-Vaya_Cisternas_Gálvez_2018}.
}
    \label{fig:EEG1020}
\end{figure}

\begin{figure}
    \centering
    \includegraphics[width=0.5\linewidth]{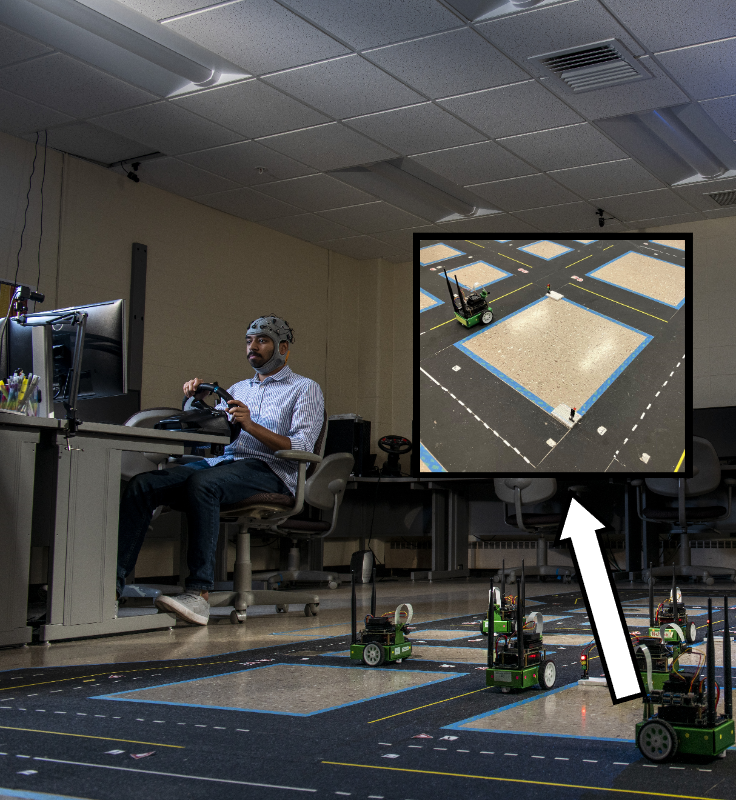}
    \caption{Photograph of the experimental setup featuring several jetbots on a portion of the testbed, a participant wearing the Enobio 20 EEG headset, the steering wheel to navigate a jetbot, and the computer monitor on which the camera feed is displayed. Insert in the photograph shows Experiment 3 set-up. Note that although multiple jetbots are shown, only the jetbot controlled by the participant is present on the testbed during the experiments in this study. Photograph credit to Micheal Pierce/Missouri S\&T.}
    \label{fig:expSetup}
\end{figure}

Three experiments are conducted to evoke and record anticipatory signals associated with the intent to actuate a brake pedal during a driving task under different conditions. These anticipatory potentials can be observed prior to the onset of the actual braking action. The conditions of the three experiments differ primarily in the imperative cues used to initiate braking and the physical and cognitive state of the participants during data collection. In all three experiments, participants are required to continuously operate a JetBot (built using Waveshare’s Jetbot AI kit and Nvidia’s 4GB Jetson-Nano) around a testbed that simulates urban streets while randomly receiving external stimuli instructing when to brake. The JetBot is controlled via a Logitech G29 Driving Force racing wheel and foot pedal setup paired with a live video feed from the JetBot on-board camera cast to a computer screen. The EEG data are collected using a Neuroelectrics ENOBIO 20 EEG headset, which is configured to use the international 10-20 standard electrode setup using 19 of the 20 channels at a data sampling rate of 500 Hz. The 20th channel serves as the common mode sense channel and is fixed to the participants right ear lobe. A schematic of the 10-20 electrode configuration and the corresponding brain lobes are shown in figure \ref{fig:EEG1020}. Data quality is monitored via real-time indicators in the collection software. The data itself is later filtered and preprocessed for model training. Figure \ref{fig:expSetup} shows a photograph of the experimental set-up.

\begin{enumerate}[label={\underline{Experiment \arabic*}:}, wide]
\setlength{\itemindent}{1ex}
\setlength{\itemsep}{2ex}
\item This experiment administers the imperative stop command via audio cues and without any additional stimulus to the driver. It is considered to be the nominal baseline dataset because data are collected from the participants in a stress-free environment. The participants are able to navigate the Jetbot with complete attention and receive the external braking stimuli via an automated, auditory countdown from 5 to 1 followed by an auditory ``stop” command, at which point the participants would depress the brake pedal with their foot. The participants would continue to keep the brake depressed, holding the Jetbot stationary, until they received another automated, auditory ``start” command. The period between start and stop commands containing the full verbal countdown, during which time EEG is recorded, is referred to as a ``trial” and 30 trials make a ``set". Detailed information about Experiment 1  can be found in  \cite{Lutes_Nadendla_Krishnamurthy_2024}. 

\item This experiment is a modified version of Experiment 1, the difference being the addition of physical fatigue and cognitive distraction elements \cite{Xu_Zhang_He_Zhao_Qi_Zhou_Zhang_Ming_2018,Almahasneh_Chooi_Kamel_Malik_2014}. As such, its procedure is mostly derived from that of Experiment 1, i.e., using audio cues to provide the braking imperative, but in this case participants are subjected to increased physical and cognitive stress before and during the EEG recording. The addition of these stressors creates conditions with the potential to induce changes to the EEG data streams, increasing the difficulty to predict the anticipatory potential, but in doing so creates a dataset more representative of data from a real-world scenario. The physical fatigue stems from requiring participants to traverse three flights of stairs (a total of 80 steps, up and down) prior to the start of every recording set. On the other hand, the cognitive distraction is administered via a sequence of text messages sent to the participants' phones at random intervals ranging from 15 to 30 seconds while they operated the JetBot and their EEG data is collected. Each message contains a simple question that the participant is required to respond to as quickly as possible without ignoring the normal driving task (i.e., requiring them to continue to navigate the Jetbot, and start and stop as instructed). An example of a question posed to the participants is ``How many is 10 divided by 2?". Note that the responses to these questions are not saved and are not part of any analysis. 

The necessity to constantly switch focus between navigating the jetbot and answering questions on a cell phone, which is not present in Experiment 1, requires increased mental effort by and increased mental workload on the participants. Because participants respond to the added stressors in different ways, in an attempt to enforce fairness, an emphasis is placed on consistency of stressor imposition rather than 
a measurement of stress sufficiency.

\begin{figure}[!t]
\centering
\includegraphics[width =0.8\textwidth]{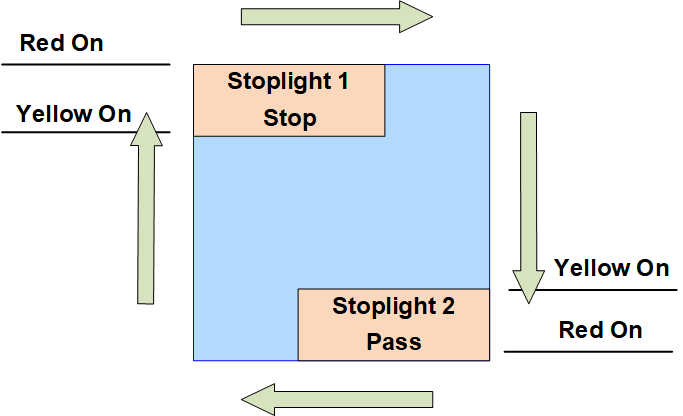}
\caption{Schematic illustrating Experiment 3 set-up (see insert in figure \ref{fig:expSetup} for a photograph of the set-up). Participants navigate the jetbot clockwise around a square track, stopping at a stoplight located at the top left corner of the track and intentionally ignoring the stoplight at the bottom right corner of the track. One revolution around the square is considered as one trial. Solid black lines with text denote the approximate locations where the noted light color would appear on the stoplight.}
\label{Figure_3}
\end{figure}

\item Compared to the two previous experiments, Experiment 3 presents the participant with a different, yet more realistic, braking imperative. In this case, participants are required to sequentially stop at, or in contrast purposefully ignore, a visual cue given in the form of a stoplight turning from yellow to red. As shown in figure \ref{Figure_3}, the domain of this experiment is constrained to a square section of the testbed with two model stoplights positioned in opposite corners of the square. The Jetbot begins at one corner of the square track and the participants navigate it around the square in a clockwise manner, approaching each stoplight sequentially. The participants are instructed to stop at the first stoplight and to continue driving without braking through the second. Upon nearing each stoplight, the participants are presented with the stoplight’s yellow light for 1.6 seconds before the stoplight switches to red. At the stoplight in which the participants stop, they remain stopped until the red light switches off at 2.15 seconds. The stoplights are activated manually by the experiment coordinator. A ``trial" for Experiment 3 is defined as one full cycle around the square track where the participants encounter both the stop and the pass-through lights. Markers are recorded within the EEG data for each time the stoplights turned yellow and red such that each trial has two sets of yellow followed by red light markers.

\end{enumerate}

The total number of participants in the three experiments are 15, 11 and 11, respectively; however, for the analysis, only 11 participants from Experiment 1 are considered for consistency among all three experiments.  Some participants participated in multiple or all experiments; however, the 11 participants chosen from Experiment 1 are not necessarily the same as the participants in the other two experiments. There are fewer participants in Experiments 2 and 3 because some of the participants from Experiment 1 were unavailable during the times when those experiments were completed, or they chose not to participate. All participants were Missouri University of Science and Technology students or professors in healthy condition with normal or corrected-to-normal vision, and with normal hearing. The three experiments received approval from the University of Missouri Institutional Review Board and all experiments were performed in accordance with relevant guidelines and regulations. Written informed consent was obtained from all participants prior to their participation. Note that because each experiment is considered separately in the statistical analysis that is explained later, it is not necessary to include the same participants in each experiment, thereby providing flexibility in recruiting participants.

\subsection{Data Preprocessing} \label{dataset_create}

In Experiment 1, each participant completes eight sets, each comprising of thirty trials, resulting in a total of 240 trials. In Experiment 2, each participant completes five sets of thirty trials each, resulting in a total of 150 trials. Only five sets are completed in Experiment 2 to accommodate participant time and availibility constraints. The experiments are carefully monitored by the authors and trials are removed if a participant fails to actuate the brake within 0.25 seconds of the stop command or if an incident occurs which might corrupt the data. Timing of the braking activity is enforced to minimize significant variations in the time period when anticipatory brain potentials are observed between trials. Datasets of both Experiments 1 and 2 are cleaned and preprocessed following a procedure outlined in \cite{Lutes_Nadendla_Krishnamurthy_2024}, which results in each trial being broken into segments between counts. The segments between counts 5 and 4, 4 and 3, 3 and 2, and 2 and 1 are given labels of 0 (negative). On the other hand, the segment between 1 and stop is given a label of 1 (positive). The remaining EEG data are not considered. The segments are padded to have the same overall length of 996 columns. With 19 channels, there are 18,924 data points for each segment. 

\begin{figure}[!t]
\centering
\includegraphics[width =\textwidth]{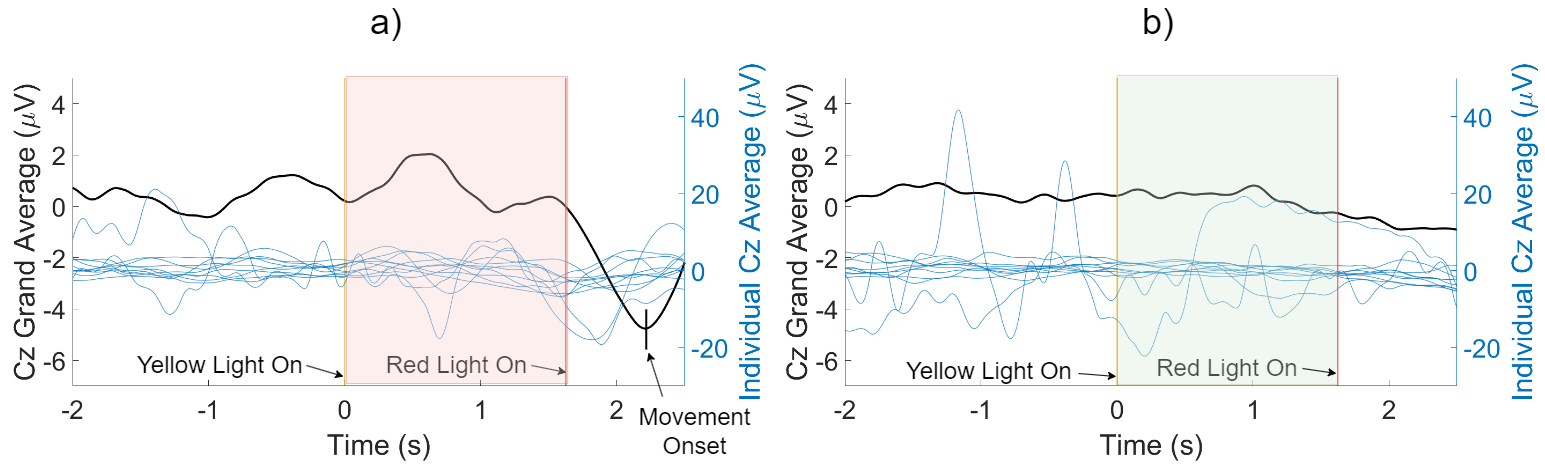}
\caption{Experiment 3 data with markers for yellow and red traffic lights. Black line is the Cz grand average of all individuals across all trials. Blue lines are Cz averages for each of the 11 individual participants across all trials. Only data collected during the time period that is shaded in red or green is used for analysis. (a) Case where participants stopped at the red light. (b) Case where participants ignored and drove through the red light without stopping.}
\label{exp3_CzGrndAve}
\end{figure}

For Experiment 3, each participant completes 3 sets of 30 trials (complete revolutions around the square track), resulting in 90 trials per participant. Similar to both Experiments 1 and 2, participants must actuate the brake pedal within a specified time limit, 0.4 seconds in this case, after red light activation or the trial is discarded. As previously noted, this automatic rejection mechanism is intended to minimize variation in the temporal location of the anticipatory brain potentials. This experiment follows the same preprocessing scheme as Experiments 1 and 2, with the exception of the segmentation strategy. Instead of five segments, only two segments are extracted from each trial: one comprising the EEG data corresponding to the traffic light in which the participant stopped and the other corresponding to the traffic light in which the participant passed through without stopping. The segments began at the yellow light marker and end at 2.15 seconds after the yellow light marker. The segment end time is determined by observing the location of the onset of movement in the stop data, evidenced by the peak of the dip in the Cz electrode (see figure \ref{fig:EEG1020}) signal grand average, and ending the segment immediately prior to this peak. The Cz grand average is calculated by averaging the Cz electrode signal from all participants and across all trials. This average along with data markers and segmentation are shown in Figure \ref{exp3_CzGrndAve}. Because the data segments for both stop and pass-without-stopping instances are the same size, 1074 data points, no additional padding is needed for this data set. 

For all three experiments, the preprocessed data are converted from floating-point data to 8-bit signed integers by quantizing the data into integer bins ranging from -127 to 128. The data quantization is necessary for compatibility with BrainChip MetaTF ML framework, which comprises three Python packages, used with the Akida processor \cite{Akida_Examples_documentation}.

\begin{table*}[!t]
\centering
\caption{Class distributions statistics of group- and individual-level model subsets. Note that because the analysis is repeated ten times for rigor, the sizes of the subsets varied from iteration to iteration when different drivers are included in the group- and individual-level model subsets for each iteration.}
\begin{tabular}{c c c c c c}
\hline \hline
\\[-2.5ex]
Experiment & \multirow{2}{*}{Subset} & \multirow{2}{*}{Driver} & Total & Positive Class & Negative Class
\\[0.5ex]
Number & & & Mean (SD) & Mean (SD) & Mean (SD)
\\[1ex]
\hline \hline
\\[-1.5ex]
\multirow{4}{*}{1} & Group & Drivers 1-8 & 4931 (318) & 986 (64) & 3944 (254)
\\[1ex]
& Individual & Driver 9 & 462 (265) & 91 (58) & 371 (207)
\\[1ex]
& Individual & Driver 10 & 620 (129) & 122 (31) & 498 (99)
\\[1ex]
& Individual & Driver 11 & 695 (136) & 141 (29) & 554 (108)
\\[1ex]
\hline
\\[-1ex]
\multirow{4}{*}{2} & Group & Drivers 1-8 & 2909 (246) & 561 (56) & 2347 (191)
\\[1ex]
& Individual & Driver 9 & 282 (130) & 53 (30) & 229 (101)
\\[1ex]
& Individual & Driver 10 & 388 (128) & 76 (27) & 313 (101)
\\[1ex]
& Individual & Driver 11 & 420 (64) & 83 (15) & 337 (50)
\\[1ex]
\hline
\\[-1ex]
\multirow{4}{*}{3} & Group & Drivers 1-8 & 948 (57) & 476 (29) & 472 (29)
\\[1ex]
& Individual & Driver 9 & 128 (29) & 64 (15) & 64 (15)
\\[1ex]
& Individual & Driver 10 & 127 (31) & 63 (15) & 64 (15)
\\[1ex]
& Individual & Driver 11 & 94 (52) & 48 (25) & 46 (27)
\\[1ex]
\hline
\\[-2ex]
\end{tabular}
\label{IndModStat}
\end{table*}

The datasets from all three experiments are split into group-level data, consisting of: i) conglomeration of eight participants' data; and ii) individual-level data, consisting of data from the remaining three participants. Thus, in total there are four subsets of data for each experiment's dataset: one group-level subset and three individual-level subsets. To obtain a more generalized set of results and prevent bias stemming from cherry-picking participants, the analysis is repeated ten times by randomly choosing different participants from the respective experiment datasets to compose the group and individual-level subsets each time, with each participant being in the individual-level subsets at least once. 

The averages and standard deviations of the total positive and negative examples within the group and individual subsets of each respective experiment are shown in table \ref{IndModStat}. Note that the driver number in the table is used merely for counting purposes and is not meant to associate an individual to the number. The datasets from Experiment 1 and Experiment 2 suffer from a class imbalance problem of approximately four negative classes for every one positive class. This is typically solved by calculating class weights for each class and incorporating these weights into the training loss function. This approach is used for training the group model with the class weights being calculated as:

\noindent
\begin{equation}
w_{c,i} = \frac{N_{D}}{2N_{c,i}}
\end{equation}

\noindent where $N_{D}$ and $N_{c,i}$ are the total number of data points in the training partition and the number of class $i$ data points in the training partition, respectively. However, this solution is not compatible with Akida edge learning, which does not utilize a typical loss function. As such, for training the individual models, the positive class samples in the individual-level subsets are duplicated three times to approximately equalize the number of positive and negative class examples. Experiment 3 dataset had approximately equal numbers of positive and negative class inputs; therefore, no additional remedy is required.

\subsection{Hardware}\label{hardware}
The neuromorphic processor used in this work is BrainChip's Akida AKD1000 NSoC. This processor is capable of performing on-chip inference as well as limited on-chip training via the Akida edge learning algorithm, which enables updating the spiking output layer's binary weights online \cite{Akida_Examples_documentation}. It is an event-based processor that is highly energy-efficient and does not require an external CPU to operate. The Akida AKD1000 processor contains 80 neural processing units, thereby enabling modeling of 1.2 million neurons and 10 billion synapses \cite{10.3389/fnins.2022.959626}. Because of the edge-learning capability, Akida AKD1000 processor can support various online learning methods such as incremental, continuous and one-shot learning at the processor level. As explained in the next section, for group-level model training, a PC with 32GB of DDR4 RAM and an Intel Xeon W-2123 3.60GHz processor with four cores and two threads per core for a total of eight threads is used.

\begin{figure}
\centering
\includegraphics[width =\textwidth]{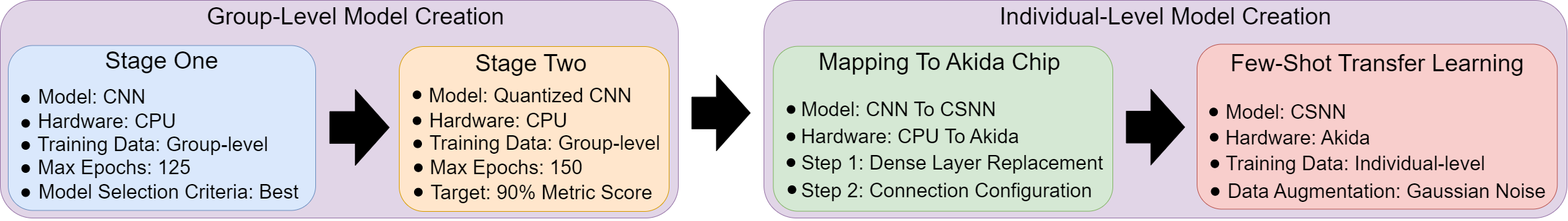}
\caption{Schematic of the few-shot transfer learning method, which incorporates a three-step approach; the first two sub-steps creates a group-level model on the CPU hardware (off-chip learning), and the third step maps this model to the Akida NsoC and implements few-shot learning on the individual-level data to create individual-level models (on-chip learning).}
\label{fig:method}
\end{figure}

\section{Method}\label{method}

\subsection{Few-Shot Transfer Learning}

The few-shot transfer learning method is designed to be compatible with the Akida AKD1000 processor. It involves the following three-step process (see figure \ref{fig:method}), which has been successfully used to solve classification problems in edge-computing applications, including weed detection in agriculture \cite{10034619} and cloud cover detection for small satellites \cite{10191569}. 

\begin{enumerate}[wide, labelwidth=!, labelindent=0pt]
\item \textbf{\emph{Group-level model creation}} - The group-level model is created off-chip in two stages. A continuous convolutional neural network (CNN), modeled using Tensorflow \cite{tensorflow2015-whitepaper}, is first created and trained on group-level data consisting of data from 8 out of 11 drivers. As shown in figure \ref{CSNN}, the CNN is a standard CNN with rectified linear unit (ReLU) activation function at the end of each convolutional layer and a linear activation function output layer with two neurons, for each of the binary classes. These activation functions are chosen based on compatibility with the MetaTF ML framework conversion library. The continuous parameter CNN is trained for 125 epochs with model parameters being saved for each epoch and the best model parameters with respect to minimizing the loss being chosen to proceed forward to the second stage involving quantization. 
The quantization process involves converting the CNN weights to discrete 8-bit integers and the activation function outputs to single-bit integers to provide the spiking behavior, effectively converting the CNN to a CSNN using the CNN2SNN package \cite{cnn2snn_source}.  The quantized network with discrete weights and activations is then fine tuned by further training on the same data to recoup any performance loss as a result of the quantization process. This training of the group-level model is terminated when the chosen classification performance metrics are greater than 90\% to prevent overfitting, which is typically less than 25 epochs, with a maximum number of training epochs of 150.

\begin{figure}[!t]
\centering
\includegraphics[width =\textwidth]{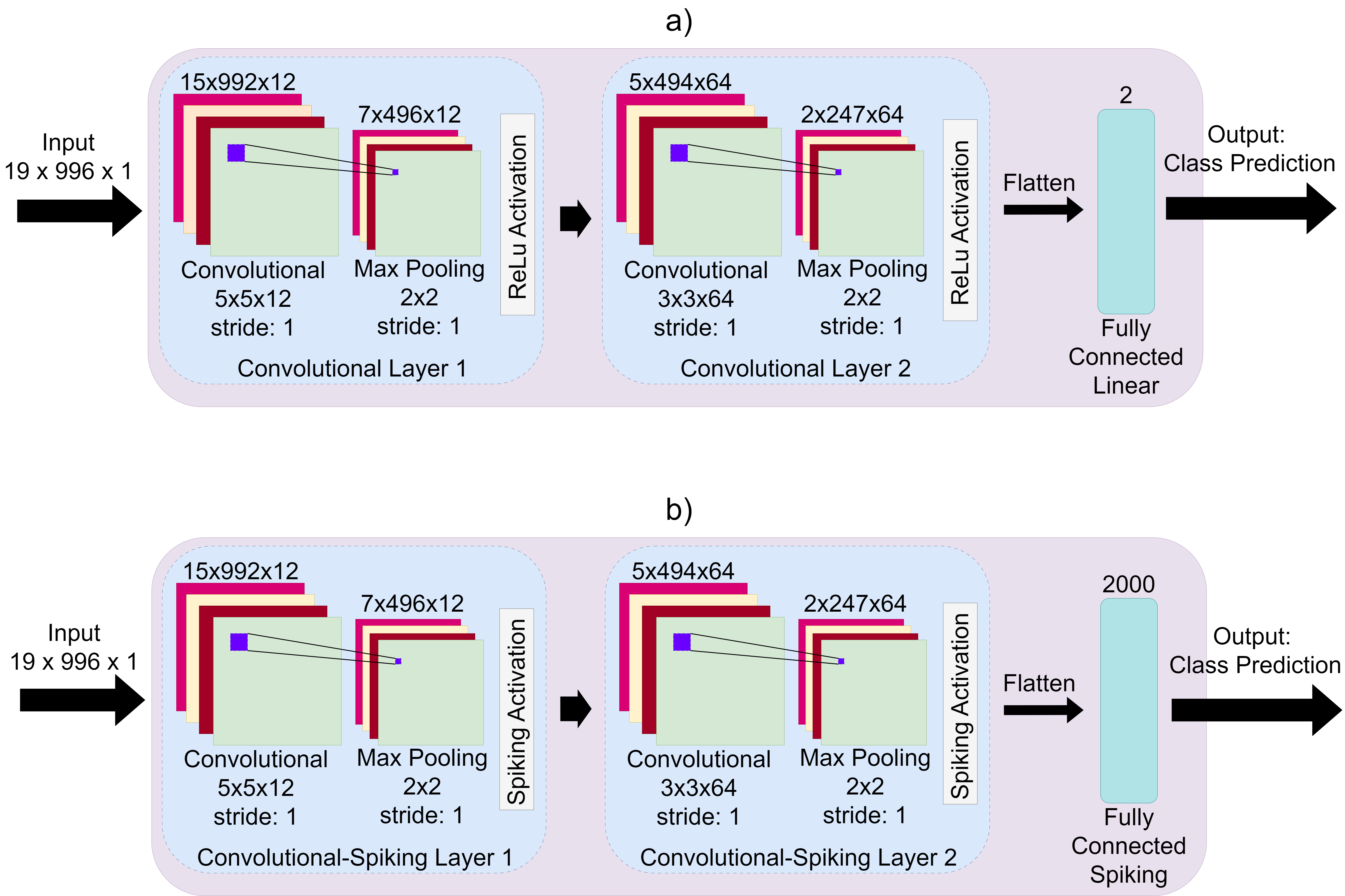}
\caption{Schematic of the neural network architectures. (a) CNN. (b) CSNN. The CNN architecture is used during the group training on the CPU hardware. It is then converted to a CSNN and mapped to the Akida NsoC for the individual-level training. Note the large increase in output neurons for the CSNN compared to the CNN. These additional neurons are required for adequate class distribution learning during the Akida unsupervised edge learning process.}
\label{CSNN}
\end{figure}

\item \textbf{\emph{Mapping the CSNN to the Akida AKD1000 processor}} – After the final group-level model parameters are determined, the model undergoes two additional steps before being mapped to the Akida processor. This first consists of replacing the previous dense output layer with a MetaTF ML framework fully connected output layer. This layer functions like a standard fully connected layer but contains binary weights and includes many more output neurons than the previous dense layer, as illustrated by the ``2000" annotation over the final layer in figure \ref{CSNN}. The extra neurons are included because of the Akida edge learning algorithm which requires there to be many neurons per class to better represent the class variability, similar to clustering algorithms where the cluster represents the distribution of data. Replacing the trained last layer with a new last layer allows the model to be customized for new data while relying on the feature extraction layers trained during the development of the group-level model. Such an approach is in-line with classical transfer learning theory \cite{Zhuang_Qi_Duan_Xi_Zhu_Zhu_Xiong_He_2021}. 

The second step is to configure the new fully connected layer by defining the number of weights that connect the previous layer to the final layer; an important parameter for the edge learning algorithm. This value determines the number of non-zero, trainable synaptic weights for each neuron. Following BrainChip's guidelines \cite{Akida_Examples_documentation}, the number of connecting weights is determined by calculating the mean of the total spiking activity emanating from the second convolutional layer (the preceding layer to the final, fully connected layer) across the individual-level subset and multiplying this value by a constant set to be equal to 1.2. With the addition of the new final layer, the CSNN is mapped to the Akida processor for creating individual-level models through few-shot learning.

\item \textbf{\emph{Few-shot on-chip learning for creation of individual-level models}} - Individual-level data from three drivers, held out from training the group-level model, are used for creating the individual-level models. Unlike training the group-level models on the CPU hardware, individual-level models are trained on the Akida processor. For each individual driver, the edge learning algorithm is used to refine the last layer parameters, with each training task starting from a new, randomly initialized final layer. Since the individual-level subsets are considerably smaller than the group-level subset, a data augmentation approach of creating additional samples by adding white noise with zero mean and one standard deviation is employed \cite{Rommel_Paillard_Moreau_Gramfort_2022,Wang_Zhong_Peng_Jiang_Liu_2018}. Because the analysis is repeated 10 times, a total of 30 individual-level models are trained per experiment for a grand total of 90 individual-level models for all three experiments. Four augmentations are created for each data sample thus increasing the size of each individual-level subset by five fold. Although the objective is for few-shot learning, individual model training is carried out for 25 epochs with the training metrics recorded for each epoch to study the effect of longer training on the classification results.

\subsection{Neural Network Details}
As shown in figure \ref{CSNN}, the initial CNN model has a layout consisting of two convolutional layers ending in ReLU activation functions followed by a final fully connected layer. The first convolutional layer has 12 filters, a kernel size of 5$\times$5 and has a max pooling layer with a pool size of 2$\times$2. The second convolutional layer has 64 filters, a kernel size of 3x3 and a max pooling layer with a pool size of 2$\times$2. All convolutional layers and max pooling layers used the default layer padding of "valid" or no padding. The final layer is a fully connected layer with linear activation and two outputs (for the binary labels) with the loss function operating on the logits output by the network. When the CNN is converted to the CSNN and mapped to the Akida processor, the weights and activations are quantized to produce spiking behavior. The final layer is also replaced with a fully connected spiking layer with 2000 neurons (1000 for each class, which is determined by experimentation based on model performance). The default values of the edge learning algorithm hyper-parameters are used. These values are: ``initial\_plasticity" = 1.0, ``learning\_competition" = 0.0, ``min\_plasticity" = 0.1 and ``plasticity\_decay" = 0.25.

\end{enumerate}

\section{Results}\label{results}

\subsection{EEG Signal Grand Averages}

\begin{figure}[!t]
\centering
\includegraphics[width =\textwidth]{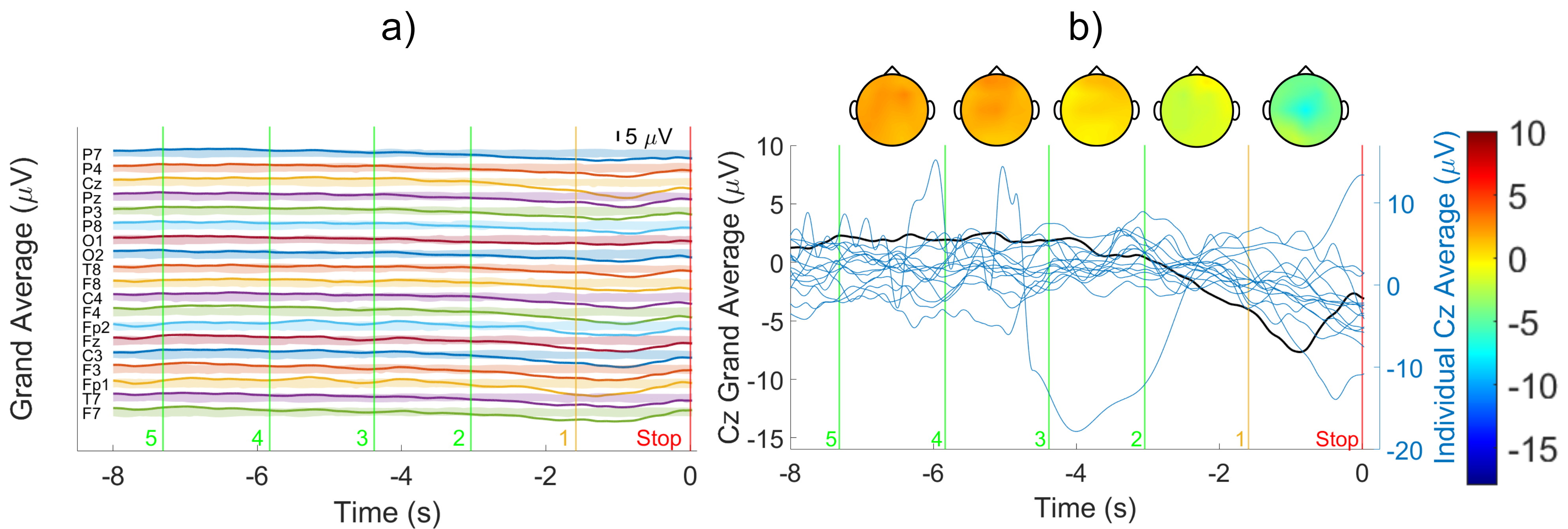}
\caption{Experiment 1 preprocessed EEG signals. (a) Channel potentials with associated countdown and “Stop” command markers and scale. Solid lines are the grand averages of all individuals across all trials. The shaded regions represent the range from minimum to maximum values across all trials (drawn at 1/40 scale of the grand average). (b) Black line is the Cz grand average signal of all individuals across all trials with scalp maps representing the grand average at the midpoint between two neighboring markers. Blue lines are Cz averages for each of the 11 individual participants across all trials. Color bar on the right displays the Cz channel potential in µV (adapted from "Convolutional spiking neural
networks for intent detection based on anticipatory brain potentials using electroencephalogram" by Nathan Lutes, Venkata Sriram Siddhardh Nadendla \& K. Krishnamurthy \cite{Lutes_Nadendla_Krishnamurthy_2024}), used under CC BY 4.0 (http://creativecommons.org/licenses/by/4.0/) - scalp maps redrawn using different scale).
} 
\label{exp1_AllGrndAve}
\end{figure}

\begin{figure}[!t]
\centering
\includegraphics[width =\textwidth]{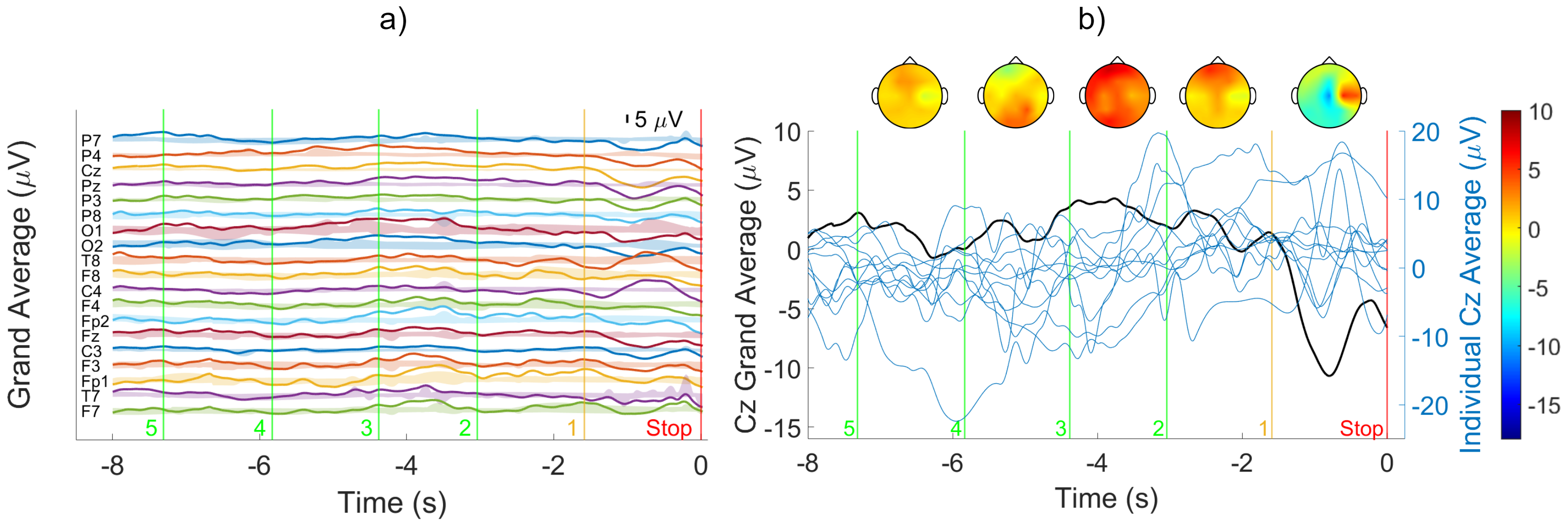}
\caption{Experiment 2 preprocessed EEG signals. (a) Channel potentials with associated countdown and “Stop” command markers and scale. Solid lines are the grand averages of all individuals across all trials. The shaded regions represent the range from minimum to maximum values across all trials (drawn at 1/300 scale.) (b) Black line is the Cz grand average signal of all individuals across all trials with scalp maps representing the grand average at the midpoint between two neighboring markers. Blue lines are Cz averages for each of the 11 individual participants across all trials. Color bar on the right displaying the Cz channel potential in µV. Grey lines represent Cz averages for each individual participant across all trials.
}
\label{exp2_AllGrndAve}
\end{figure}

\begin{figure}[!t]
\centering
\includegraphics[width =\textwidth]{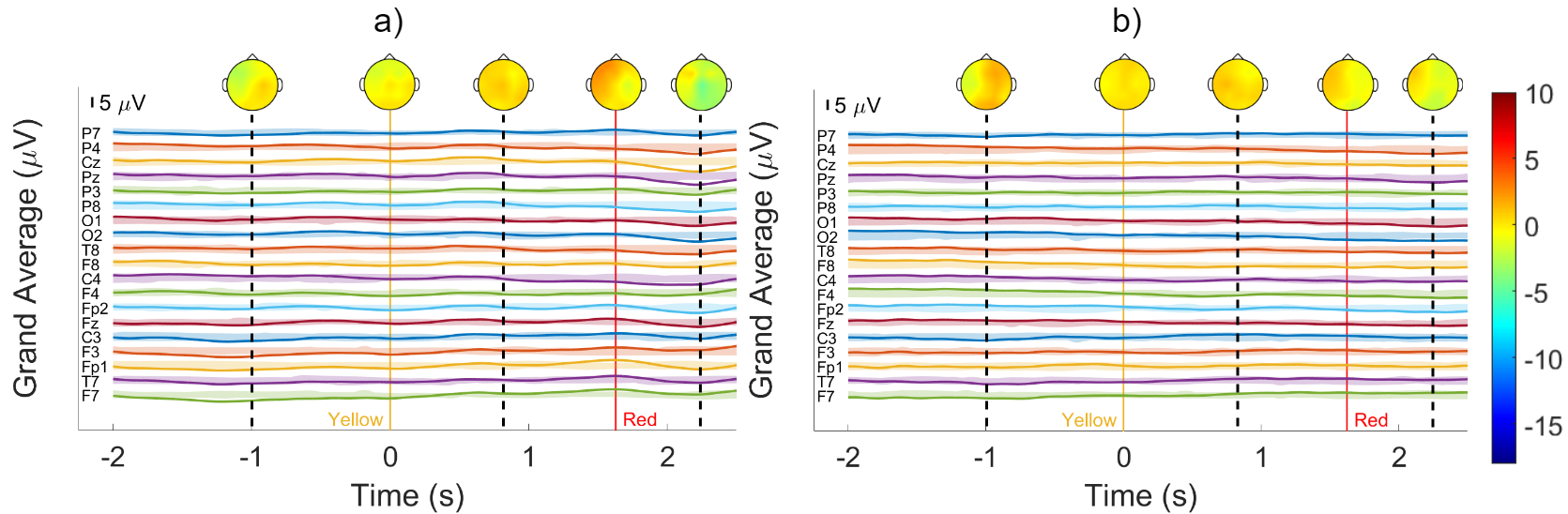}
\caption{Experiment 3 preprocessed EEG signals. (a) Channel potentials for stopping at red light with markers for yellow and red lights, and scale. (b) Channel potentials for driving through without stopping at the red light with markers for yellow and red lights, and scale. For both (a) and (b), Solid lines are the grand averages of all individuals across all trials. The shaded regions
represent the range from minimum to maximum values across all trials (drawn at 1/40 scale of the grand average). Also, scalp maps represent the grand average at the marker location or at the temporal location denoted by the dashed lines. The color bar on the right displays the channel potential in µV.}
\label{exp3_AllGrndAve}
\end{figure}

The grand averages calculated by averaging the EEG signals from all 11 participants across all trials, individual averages calculated by averaging the EEG signals for each of the 11 individuals across all trials, and the associated EEG scalp plots are shown in figures \ref{exp1_AllGrndAve}, \ref{exp2_AllGrndAve} and \ref{exp3_AllGrndAve} for Experiments 1, 2 and 3, respectively. In all three figures, the EEG signals are staggered for better visualization and the scalp plots have the same color bar range. Several interesting observations can be made. 

A sharp negative variation can be observed in the Cz grand average in all three figures. In Experiments 1 and 2, this begins around the 2 marker and continues until about 1 second before the stop marker. In Experiment 3, this can be observed in figure \ref{exp3_AllGrndAve}(a), at the red light activation marker and continues for about 0.6 seconds after the marker. However, this is noticeably absent in figure \ref{exp3_AllGrndAve}(b), which shows the EEG grand averages of the instances where the participant did not stop. This negative variation, referred to as a contingent negative variation (CNV), is indicative of movement intent \cite{khaliliardali_chavarriaga_gheorghe_millan_2015,khaliliardali_chavarriaga_zhang_gheorghe_perdikis_millan_2019,garipelli_chavarriaga_del_r_millan_2013} and can be seen in several areas of the brain across all three experiments, but most notably occurs in the centro-medial region. The primarily centralized location of the phenomenon is the motivation for using the Cz grand average as the basis for the segmentation boundaries as illustrated in figure \ref{exp3_CzGrndAve}. Indeed, an ablation study performed in \cite{Lutes_Nadendla_Krishnamurthy_2024} confirmed that only a few channels located in the centro-medial region are necessary for satisfactory prediction of the CNV signal.

The effects of physical fatigue and cognitive distraction can be observed by comparing Experiments 1 and 2 scalp plots (figures \ref{exp1_AllGrndAve}(b) and \ref{exp2_AllGrndAve}(b)). In general, the EEG grand averages for Experiment 2 appear to have much more variability than Experiment 1 indicating dynamic brain activity. Experiment 2 shows activity in the O1 and O2 electrodes over the occipital lobe and P4 over the parietal lobe between the 4 and 3 markers. While it is difficult to pinpoint the exact cause of the activation because of volume conduction, increased activity in these electrodes indicate visual and cognitive processing, respectively \cite{eeg_chan_assoc_map,Walker_2009}. Experiment 2 also shows a period of high positive activity across many of the brain regions between the 3 and 2 markers. Between the 2 and 1 marker, the F7 and F8 electrodes over the prefrontal cortex and the Fp1, Fp2, F7, F3, Fz, F4 electrodes over the frontal lobe are active indicating cognitive and executive functions, respectively. The scalp plot before the stop command in figure \ref{exp2_AllGrndAve}(b) shows a small region on the right lateral part of the brain that has the opposite effect of the CNV, instead increasing to large positive values. This is because of the C4 and T8 electrodes as can be seen in figure \ref{exp2_AllGrndAve}(a). Increased activity in the C4 electrode over the central region arises from executing hand movements. Participants are instructed to rapidly respond to the cognitive distraction questions on their cell phones, thus requiring them to quickly remove their hands from the wheel, respond and then re-engage the steering wheel. The T8 electrode over the right temporal lobe shows increased activity because of auditory processing and memory recall that occurs while responding to the cognitive distraction questions. The increased brain activity levels and steeper decline in the CNV is the effect of physical fatigue and cognitive distraction throughout the countdown leading to a much shorter reaction time for participants in Experiment 2 who are less prepared for the stop command than those in Experiment 1. Interestingly, the magnitude in the Cz grand average CNV signal is much larger for Experiment 2 than for Experiment 1. This is also because of the reduced reaction time for distracted participants causing the brain to make rapid preparations for the upcoming braking action. Although it is not possible to disentangle the effects of physical fatigue and cognitive distraction, it is posit that the higher brain activity is the result of cognitive distraction.

In figure \ref{exp3_AllGrndAve}, besides the CNV in figure \ref{exp3_AllGrndAve}(a), notable differences between figures \ref{exp3_AllGrndAve}(a) and \ref{exp3_AllGrndAve}(b) occur one second prior to the yellow light marker and at the red light marker. Prior to the yellow light marker in figure \ref{exp3_AllGrndAve}(b), the Fp2 and F4 electrodes over the frontal lobe and the O1 and O2 electrodes over the occipital lobe have higher activity, indicating restraint on seeing the stoplight as per the experiment instructions. At the red light marker in figure \ref{exp3_AllGrndAve}(a), there is higher activity on the left and front sides of the brain compared to the instance where participants purposefully neglect to stop. Higher activity is seen in the F7 electrode over the prefrontal cortex, F3, Fp1, Fp2 and Fz over the frontal lobe, T7 over the temporal lobe and C3 over the central region indicating cognitive, motor, memory and executive task activities associated with increased attention, recollection of the experiment instructions to stop and action planning.

\subsection{Analysis Results: Performance}

\begin{figure}
    \centering
    \includegraphics[width=\linewidth]{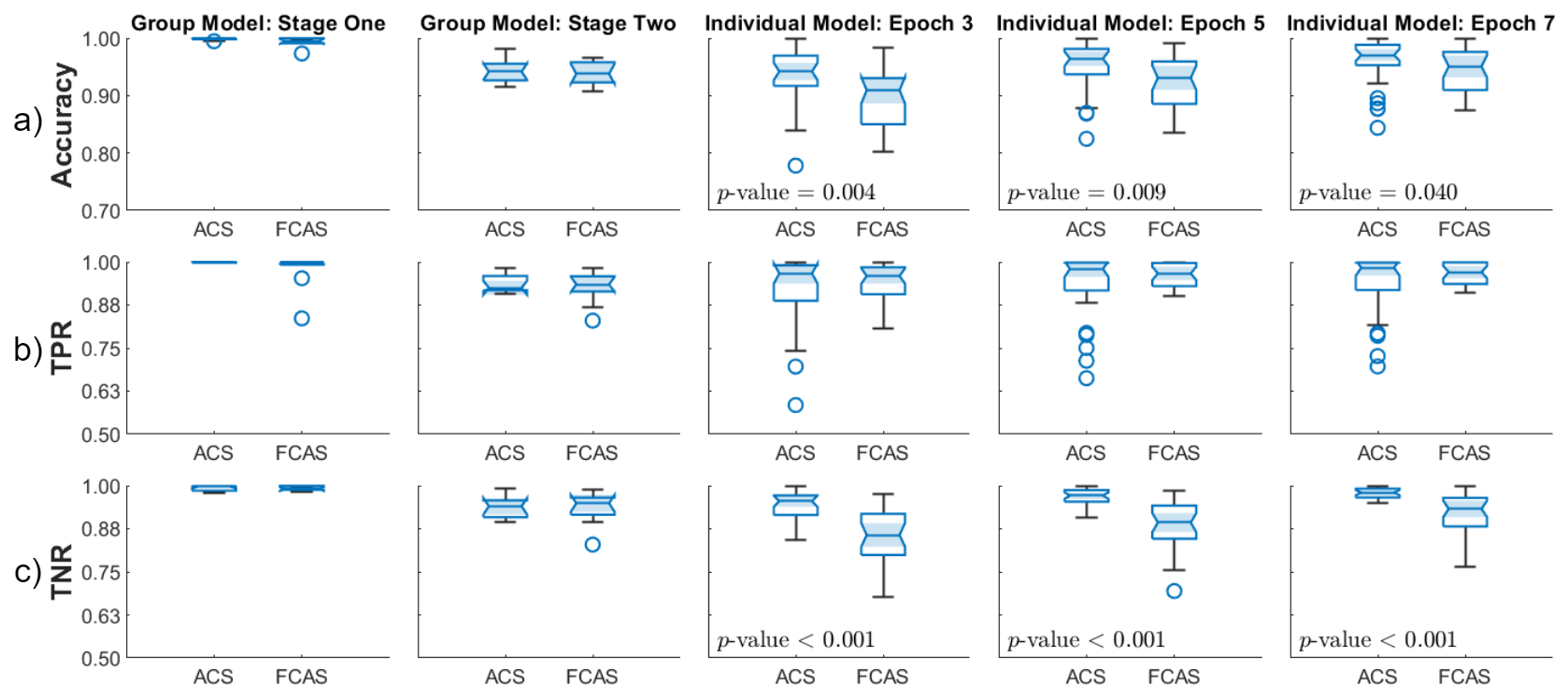}
    \caption{Experiment 1 training results for group- and individual-level models. (a) Accuracy. (b) True positive rate (TPR). (c) True negative rate (TNR). On each notched box, the central mark shows the median, bottom and top edges show the 25th and 75th percentiles, respectively, and the whiskers extend to the most extreme data points that are not considered as outliers, which are defined as 1.5$\times$ the interquartile range away from the top or bottom of the box. Each subplot shows a comparison of the results for all-channel study (ACS) and five-channel ablation study (FCAS) when the analysis is repeated 10 times. Non-overlapping shaded notched regions in a subplot implies the median values are statistically different at the 5\% level for the particular subplot case. In such cases, the \textit{p}-value is shown in the subplot.}
    \label{fig:exp1_box_plots}
\end{figure}

\begin{figure}
    \centering
    \includegraphics[width=\linewidth]{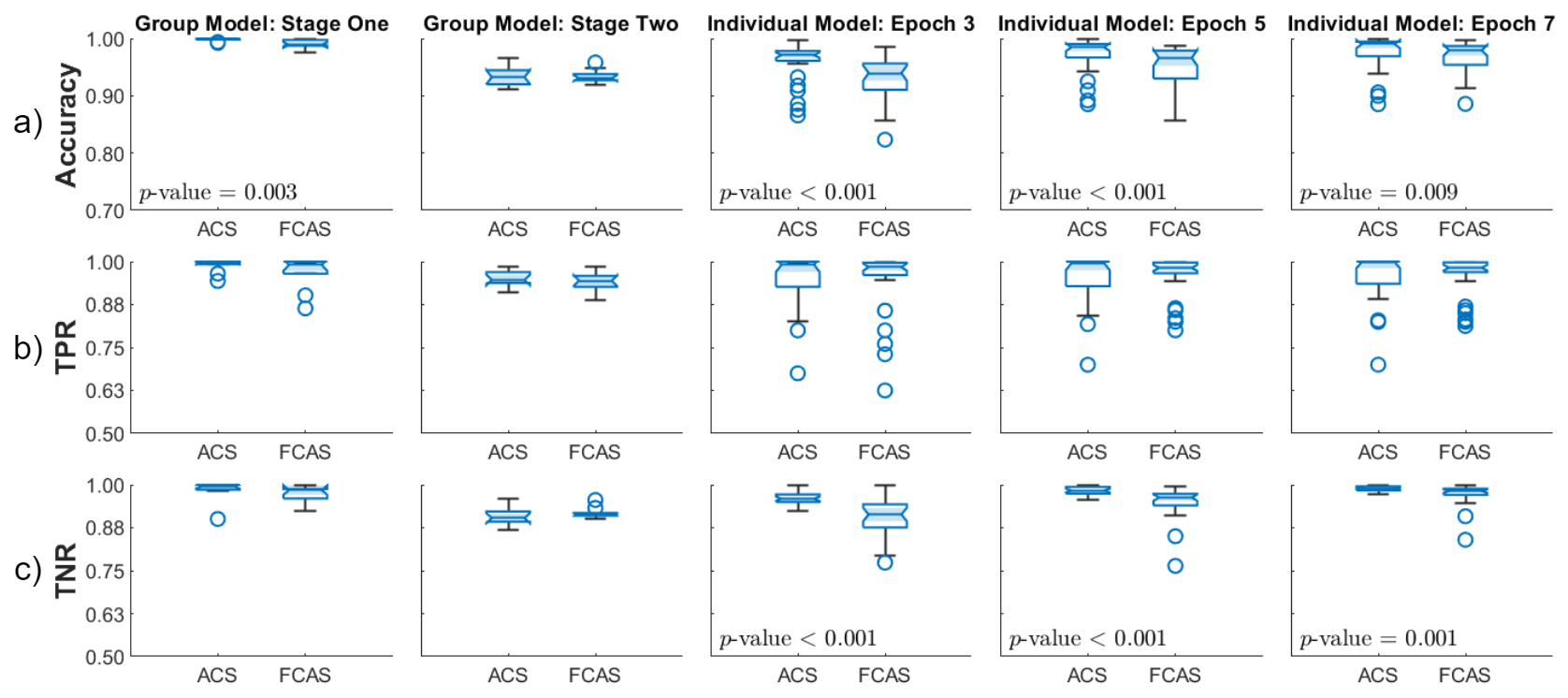}
    \caption{Experiment 2 training results for group- and individual-level models. (a) Accuracy. (b) True positive rate (TPR). (c) True negative rate (TNR). On each notched box, the central mark shows the median, bottom and top edges show the 25th and 75th percentiles, respectively, and the whiskers extend to the most extreme data points that are not considered as outliers, which are defined as 1.5$\times$ the interquartile range away from the top or bottom of the box. Each subplot shows a comparison of the results for all-channel study (ACS) and five-channel ablation study (FCAS) when the analysis is repeated 10 times. Non-overlapping shaded notched regions in a subplot implies the median values are statistically different at the 5\% level for the particular subplot case. In such cases, the \textit{p}-value is shown in the subplot.}
    \label{fig:exp2_box_plots}
\end{figure}

\begin{figure}
    \centering
    \includegraphics[width=\linewidth]{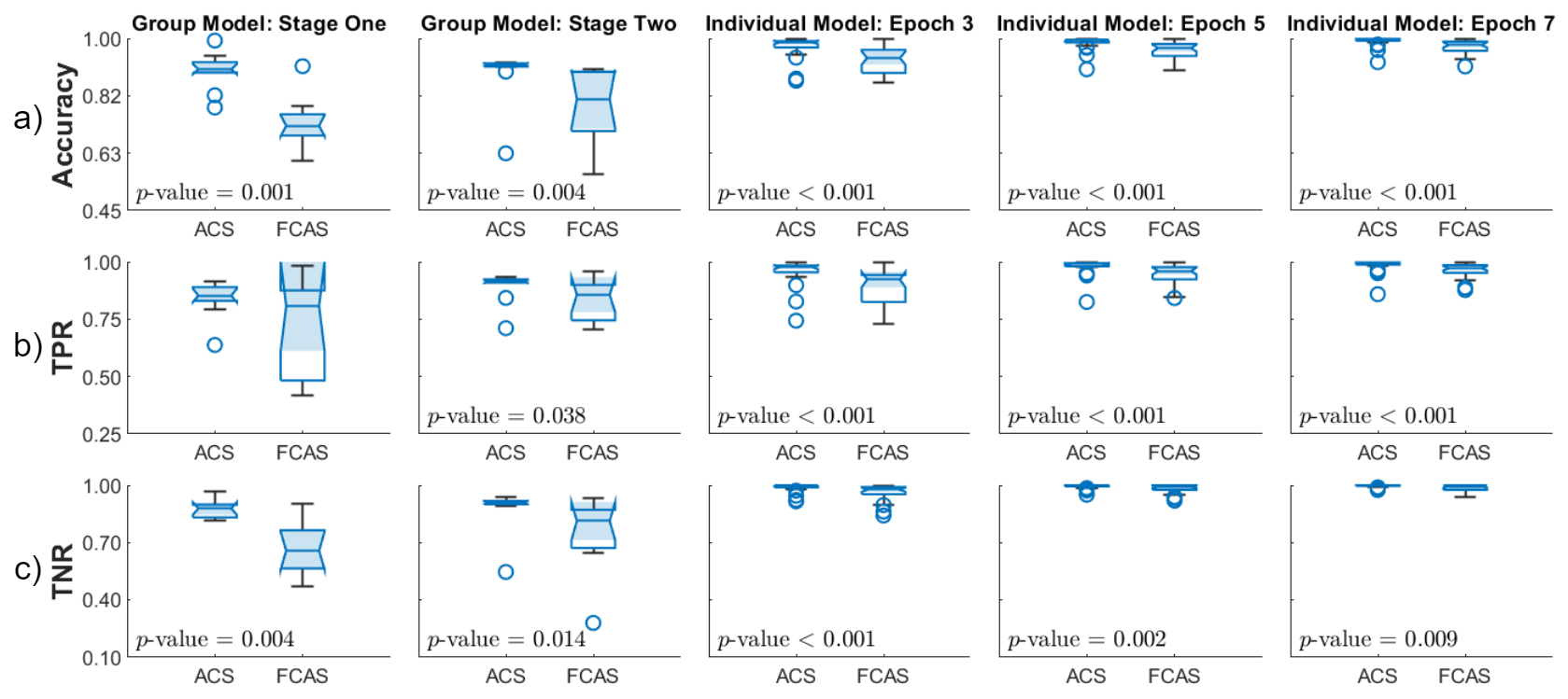}
    \caption{Experiment 3 training results for group- and individual-level models. (a) Accuracy. (b) True positive rate (TPR). (c) True negative rate (TNR). On each notched box, the central mark shows the median, bottom and top edges show the 25th and 75th percentiles, respectively, and the whiskers extend to the most extreme data points that are not considered as outliers, which are defined as 1.5$\times$ the interquartile range away from the top or bottom of the box. Each subplot shows a comparison of the results for all-channel study (ACS) and five-channel ablation study (FCAS) when the analysis is repeated 10 times. Non-overlapping shaded notched regions in a subplot implies the median values are statistically different at the 5\% level for the particular subplot case. In such cases, the \textit{p}-value is shown in the subplot.}
    \label{fig:exp3_box_plots}
\end{figure}

\begin{figure}[!t]
    \centering
    \includegraphics[width = \textwidth]{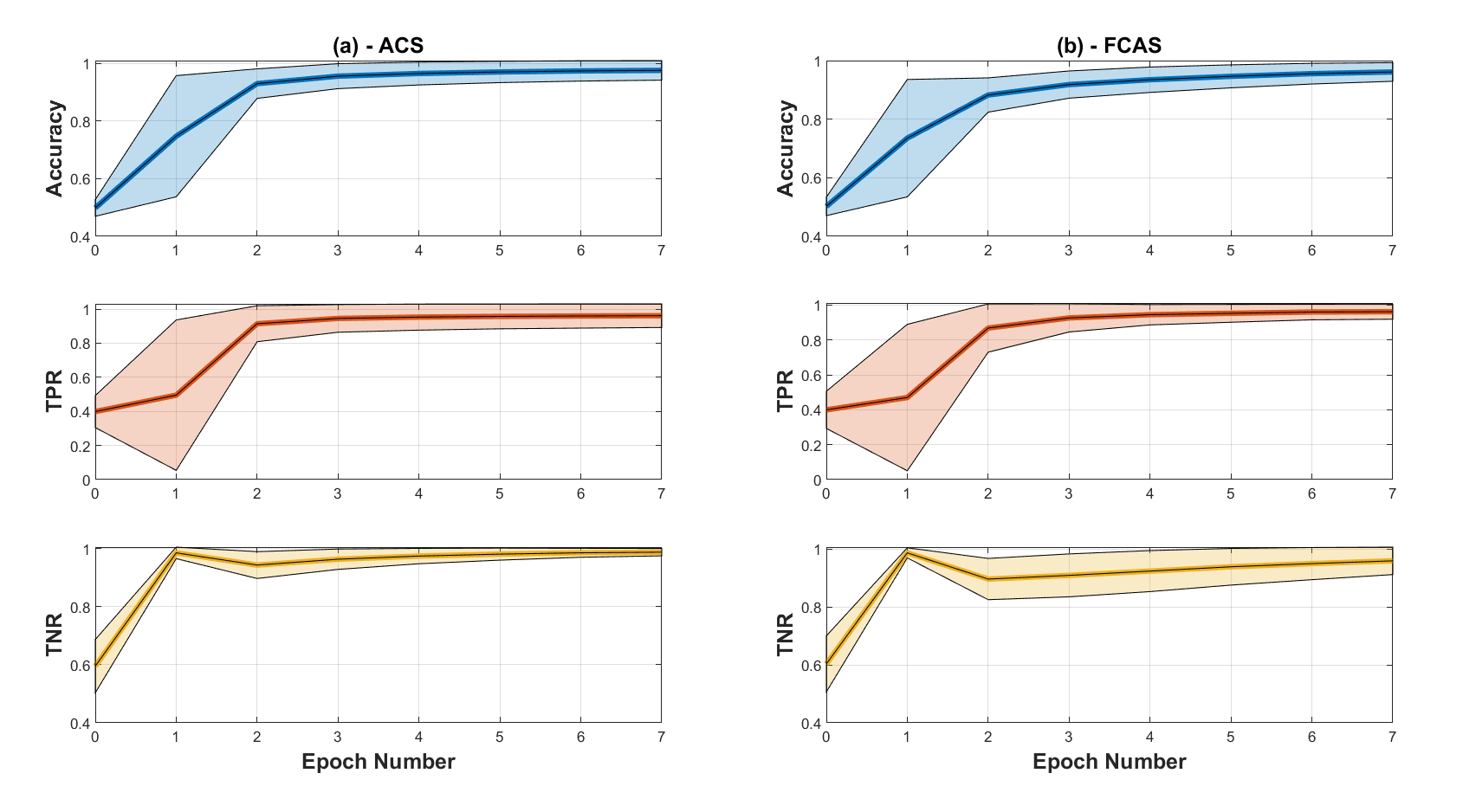}
    \caption{Performance of individual-level models as a function of transfer learning shots (epochs) for learning competition = 0 (default). (a) Average metric score with one standard deviation shaded region for accuracy, true positive rate (TPR) and true negative rate (TNR), respectively, for all-channel study (ACS). (b) Average metric score with one standard deviation shaded region for accuracy, TPR and TNR, respectively, for Five-channel Study (FCAS).}
    \label{metrics_v_epochs}
\end{figure}

Results are presented for two different studies to evaluate the efficacy of the proposed methodology: i) a baseline all-channel study (ACS) utilizing all 19 channels; and ii) a five-channel ablation study (FCAS) using only the Cz, Pz, C3, C4 and Fz centro-medial channels as done in \cite{Lutes_Nadendla_Krishnamurthy_2024}. To implement the ablation study, unused data channels are replaced with null values to allow usage of the same architecture configuration. To determine if the median values obtained from the ACS and FCAS are statistically different within each of the three experiment, two-sided Wilcoxon rank sum tests are conducted. Figures \ref{fig:exp1_box_plots}, \ref{fig:exp2_box_plots} and \ref{fig:exp3_box_plots} show results for the three experiments for the first two sub-steps when group-level models are created and at select intervals during the creation of the individual-level models (see figure \ref{fig:method}). Each figure shows notched box plots representing the distribution of median values for each performance metric, i.e., classification accuracy, true positive rate (TPR) and true negative rate (TNR), for both studies when the analysis is repeated 10 times. The \textit{p}-values obtained are also shown if there is no overlap in the notches of two box plots in each subplot, indicating that the ACS and FCAS are statistically different at the 5\% significance level. Because the method presented in this study is a few-shot learning method, only the training metrics are reported, consistent with previous literature \cite{scherr_stockl_maass_2020}.

Figure \ref{metrics_v_epochs} shows the average of the metric scores as training of the individual models in all three experiments progresses across the training epochs. The metric score averages are shown as continuous lines and the shaded regions correspond to one standard deviation. Note that the values reported at epoch ``0" in the figure are the averages and standard deviations of the performance metric scores obtained from a preliminary inference using individual models with initial, untrained final layers before any edge learning occurs. The results shown in figure \ref{metrics_v_epochs} are obtained using the zero default value of the ``learning\_competition" parameter in the edge learning algorithm. Results similar to figure \ref{metrics_v_epochs} are obtained with other values of ``learning\_competition". These results are shown in figures S1 - S3 in the supplementary material.

\subsection{Analysis Results: Power and Latency}

\begin{table*}[t!]
\centering
\caption{Power consumption and inference time comparison between central processing unit (CPU) and Akida NSoC hardware for individual-level models. CNN - convolutional neural network, CSNN - convolutional spiking neural network, SD - standard deviation, ACS - all-channel study, FCAS - five-channel ablation study, PR - percent reduction, LR - latency ratio.\\
}
\resizebox{\textwidth}{!}
{
\begin{tabular}{c c c c c c c c c c c}
\hline 
\\[-2ex]
& & & \multicolumn{4}{c}{Power (W)} & \multicolumn{4}{c}{Inference Time (s)}
\\[0.75ex]
\hline\hline
\\[-1.75ex]
\multirow{2}{*}{Experiment} & \multirow{2}{*}{Model} & \multirow{2}{*}{Hardware} &
\multicolumn{2}{c}{ACS} & \multicolumn{2}{c}{FCAS} &
\multicolumn{2}{c}{ACS} & \multicolumn{2}{c}{FCAS}
\\[0.5ex]
& & & Mean (SD) & PR & Mean (SD) & PR & Mean (SD) & LR & Mean (SD) & LR
\\[0.75ex]
\hline\hline
\\[-1.5ex]
\multirow{2}{*}{1} & CNN  & CPU &  43.31 (2.35) & - & 42.06 (3.79) & - & 4.49 (3.30) & - & 5.63 (2.71) & -
\\[1ex]
& CSNN & Akida & 0.875 (.001) & 97.98 & 0.874 (0.0001) & 97.92 & 5.56 (4.15) & 1.24 & 5.36 (2.72) & 0.95
\\[1ex]
\hline
\\[-2ex]
\multirow{2}{*}{2} & CNN  & CPU & 43.1 (0.65) & - & 41.8 (1.1) & - & 3.67 (1.37) & - & 3.17 (1.97) & -
\\[1ex]
& CSNN & Akida & 0.872 (0.001) & 97.97 & 0.875 (0.001) & 97.91 & 4.80 (1.94) & 1.30 & 2.76 (1.76) & 0.87
\\[1ex]
\hline
\\[-2ex]
\multirow{2}{*}{3} & CNN  & CPU & 39.62 (5.35) & - & 37.8 (2.32) & - & 0.72 (0.29) & - & 0.87 (0.34) & - 
\\[1ex]
& CSNN & Akida & 0.874 (0.0002) & 97.79 & 0.874 (0.0001) & 97.69 & 0.94 (0.36) & 1.30 & 0.72 (0.26) & 0.83
\\[1ex]
\hline
\\[-2ex]
\end{tabular}
}
\label{PowerTable}
\end{table*}

Table \ref{PowerTable}
show the means and standard deviations of the power consumed and inference latency for all individual-level subsets across all 10 tests, for both the ACS and FCAS, subdivided by experiment. For this evaluation, each of the three held out individual-level subsets for a given test are passed as a single batch to the quantized group-model CNN on the CPU and the MetaTF ML framework based CSNN on the Akida processor and the time required and power consumed for inference are recorded for comparison. At this time, the MetaTF API only supports power consumption at an inference level and does not support measurement of the network training-level power consumption. The CPU power consumed is calculated by recording the power required for inference and then subtracting the power consumed when code execution is suspended. This is intended to provide a fairer comparison with the Akida processor by removing any power consumed by background processes such as the operating system from the inference power measurement. The CPU power was measured using the \emph{PyJoules} Python package. This package monitors the CPU energy consumed while running Python code by using the "Running Average Power Limit" (RAPL) technology that is standard on Intel CPUs. The power consumed is computed by dividing the energy usage reported for a given computation (e.g. Python script) by the duration of the computation. The percent reduction ($PR$) in the power consumed and the latency ratio ($LR$) are calculated as: 

\begin{flushright}
$ PR = 100 \times  \left( \displaystyle \frac{CPU_{power} - Akida_{power}}{CPU_{power}} \right) $\hspace{5cm}(2)$ $\\
\vspace{10pt}
$LR = \displaystyle \frac{Akida_{inference\: time}}{CPU_{inference\:time}}$ $\hspace{6cm}(3)$

\end{flushright}

\noindent Note that a latency ratio greater than 1 indicates an increased latency for Akida.

Due to the different amounts of data collected for each experiment and the data quality checks administered as previously discussed, the total number of data points differ for each individual-level subset (see table \ref{IndModStat}). Therefore, the inference power consumed and inference latency for each individual-level subset are different. Because of these differences, the data are expressed as means and standard deviations.

\section{Discussion}\label{discussion}

\subsection{Performance}

Several observations can be made considering the results for Experiment 1 shown in figure \ref{fig:exp1_box_plots}. For both the ACS and FCAS, stage one group-level training results for all three classification performance metrics have tight distributions with all values above 90\% except for a few values for FCAS TPR. Stage two group-model training shows greater distributions, but this is expected because of the stopping criteria aimed at reducing over-fitting to the quantized group model. Three observations could be made examining the individual-level model box plots. First, the majority of values in the ACS are at or above 90\% score for all metrics, even at only 3 shots (epochs) with the median and 1st and 3rd quartile values improving with more shots. The FCAS shows, in general, larger distributions, especially in the TNR metric; however, it displays higher compactness in TPR than the ACS. Second, except for TPR, ACS and FCAS results are significantly different as indicated by the non-overlapping notched regions in the box plots using a 5\% significance level. This is in contrast to the group-level model. Therefore, for better results, FCAS could be used for group-level models with reduced computational burden, whereas ACS would be necessary for individual-level models. Finally, as expected, all training results improve when training is conducted over more epochs.

Results for Experiment 2 are presented in Figure \ref{fig:exp2_box_plots}. The group model training results are similar to those obtained for Experiment 1 albeit with greater variability and slightly lower median and 1st and 3rd quartile values. This is expected because of the increased brain activity due to the physical and cognitive distraction noted in figure \ref{exp2_AllGrndAve}. The trend of results for training individual-level models are also on par with Experiment 1 results, indicating robustness of the Akida edge learning algorithm even in the presence of physical and cognitive distraction.

Figure \ref{fig:exp3_box_plots} shows results for Experiment 3 where participants stop at the traffic light or intentionally ignore the stoplight. Except for the TPR at stage one of creating group-level models, all Experiment 3 results show a significant difference between the ACS and FCAS. This is not surprising and to be expected because Experiment 3 includes a visual stimulus in monitoring the stoplight for a braking imperative. Visual information is processed in the occipital lobe, where O1 and O2 electrodes are placed; however, these channels are not included in the ablation study that focused on the centro-medial region. Thus, critical information from these channels are missing in the ablation study data.

It is apparent from the results showed in figure \ref{metrics_v_epochs} that, in general, the individual models deployed on the Akida processor adapt to each individual driver's data quickly, only requiring two epochs on average to achieve at least 90\% performance metric score when training with data from all channels and requiring only three epochs on average when training with five channels. Although the accuracy performance metric and TPR standard deviations are higher after the first training epoch, they reduce significantly after the second training epoch and beyond in both ACS and FCAS. This demonstrates an overarching consistency in the training results in just a few shots.

\subsection{Power and Latency}

Across all three experiments, both ACS and FCAS show a substantial reduction in the inference power consumption using the Akida processor compared to the CPU hardware, achieving percent reductions greater than 97\%. Interestingly, the Akida processor has a slightly larger (about 1.3$\times$) latency compared to the CPU hardware when all channels are used, but a smaller (about 0.9$\times$) latency only five channels are used in FCAS. The latency results of the ACS are not altogether surprising as the Intel Xeon CPU is a very powerful CPU capable of making a large number of computations in a short amount of time; however, the superior latency of the Akida processor in the FCAS is indicative of the competitiveness of neuromorphic hardware with cutting edge von-Neumann hardware albeit at a much lower power requirement. This corroborates with previous studies comparing neuromorphic and von-Neumann paradigms \cite{10191569, Ceolini_Frenkel_Shrestha_Taverni_Khacef_Payvand_Donati_2020}. The Akida processor achieves this through its sparse, asynchronous computing regime.

\subsection{Limitations}

While the performance of the method on the data presented shows promise, this is not achieved without limitations. First, the mechanization of the method is intrinsically coupled with BrainChip's proprietary Akida NSoC and MetaTF API, including their proprietary edge learning algorithm. Thus, adaptations of the method to other neuromorphic hardware may not be straightforward and could require modification to the methodology or it may provide less than satisfactory results. Second, the method requires prior training and maintenance of a group-level model before individual-level models can be created. This, in turn, carries over some of the drawbacks of previous, purely group-level approaches in that a large dataset must be acquired, and time and resources must be spent creating the group-level model. Lastly, the method, as it is presented in this work, is not truly "online". For actual deployment, individual-level data must be collected for every driver using a contrived data acquisition scheme and the model trained on this data before implementing and continuing to train during actual driving scenarios. This data acquisition scheme must be defined appropriately such that enough examples of different braking and normal driving conditions are present to adequately represent the driver's individual response. Nonetheless, this work presents a novel application of a methodology able to create individual-level models at the hardware level for EEG-ADAS related tasks. As such, at this time, there does not exist an available benchmark with which to compare the performance results presented.

\section{Conclusion}\label{conclusion}

The results show that the methodology presented was effective to develop individual-level models deployed on a state-of-the-art neuromorphic processor with predictive abilities for ADAS relevant tasks, specifically braking intent detection. This study explored a novel application of deep SNNs to the field of ADAS using a neuromorphic processor by creating and validating individual-level braking intent classification models with data from three experiments involving pseudo-realistic conditions. These conditions included cognitive atrophy through physical fatigue and real-time distraction and providing braking imperatives via commonly encountered visual stimulus of traffic lights. The method presented demonstrates that individual-level models could be quickly created with a small amount of data, achieving greater than 90\% scores across all three classification performance metrics in a few shots (three epochs) on average for both the ACS and FCAS. This demonstrated the efficacy of the method for different participants operating under non-ideal conditions and using realistic driving cues and further suggests that a reduced data acquisition scheme might be feasible in the field. Furthermore, the applicability to energy-constrained systems was demonstrated through comparison of the inference power consumed with a very powerful CPU in which the Akida processor offered power savings of 97\% or greater. The Akida processor was also shown to be competitive in inference latency compared to the CPU. Future work could include implementation of the method presented on a larger number of participants, other neuromorphic hardware, different driving scenarios, and in real-world scenarios where individual-level models are created by refining previously developed group-level models in real time.

\section*{Data Availability}\label{data_avail}
\noindent The datasets used and/or analyzed during the current study are available from the corresponding author on reasonable request.

\section*{Code Availability}\label{code_avail}
\noindent The Python code written to obtain the results will be publicly available at \url{https://github.com/sid-nadendla/TL-Akida} upon publication of the manuscript.

\section*{Author contributions}
 N.L, V.S.S.N. and K.K. conceptualized and designed the method, performed analysis and prepared the manuscript. N.L. collected and processed the EEG data, and developed the software code to generate the results.

\section*{Conflict of interest}
The authors declare that they have no conflict of interest.

\section*{ORCID iDs}

\noindent Nathan Lutes \url{https://orcid.org/0000-0003-0512-5809}

\noindent Venkata Sriram Siddhardh Nadendla \url{https://orcid.org/0000-0002-1289-7922}

\noindent K. Krishnamurthy \url{https://orcid.org/0000-0002-0219-6153}

\section*{References}
\bibliography{main_JEng_v7_rev1.2_clean}

\end{document}